\documentclass[runningheads]{llncs}
\usepackage[T1]{fontenc}
\usepackage{graphicx}
\usepackage{hyperref}
\usepackage{color}
\usepackage{marvosym}
\usepackage{subcaption}

\usepackage{amsmath,amssymb}
\makeatletter
\def\@fnsymbol#1{(\Letter)}
\makeatother
\begin{document}

\title{DiffExplainer: Unveiling Black Box Models Via Counterfactual Generation}
\author{Yingying Fang\inst{1}\and 
Shuang Wu\inst{2}\and 
Zihao Jin\inst{3}\and 
Shiyi Wang\inst{4} \and 
Caiwen Xu\inst{4} \and \\
Simon Walsh\inst{1}$^{\dag}$ \and 
Guang Yang\inst{1,4,5,6}$^{\dag}$} 
\authorrunning{Y. Fang et al.}

\renewcommand{\thefootnote}{}
\footnotetext{$^\dag$ S. Walsh and G. Yang are co-senior authors.}

\institute{National Heart and Lung Institute, Imperial College London, London, UK \and
Black Sesame Technologies, Singapore, Singapore \and
Department of Metabolism, Digestion and Reproduction, Imperial College London, London, UK \and
Bioengineering Department and Imperial-X, Imperial College London, London, UK 
\and
Cardiovascular Research Centre, Royal Brompton Hospital, London, UK
\and
School of Biomedical Engineering \& Imaging Sciences, King's College London, London, UK
\\
\email{\{s.walsh;g.yang\}@imperial.ac.uk} 
\\
}

\maketitle

\begin{abstract}
In the field of medical imaging, particularly in tasks related to early disease detection and prognosis, understanding the reasoning behind AI model predictions is imperative for assessing their reliability. Conventional explanation methods encounter challenges in identifying decisive features in medical image classifications, especially when discriminative features are subtle or not immediately evident. To address this limitation, we propose an agent model capable of generating counterfactual images that prompt different decisions when plugged into a black box model. By employing this agent model, we can uncover influential image patterns that impact the black model's final predictions. Through our methodology, we efficiently identify features that influence decisions of the deep black box. We validated our approach in the rigorous domain of medical prognosis tasks, showcasing its efficacy and potential to enhance the reliability of deep learning models in medical image classification compared to existing interpretation methods. The code will be publicly available at \url{https://github.com/ayanglab/DiffExplainer}.
\end{abstract}

\setcounter{footnote}{0}

\section{Introduction}

In this age where deep learning-based models are increasingly predominant, it is of paramount importance for users to comprehend the rationale behind the decisions made by these models, especially when these models are engaged in high stake applications such as AI-based diagnosis and prognosis. Over the past decade, Explainable AI (XAI) methods for medical image analysis have been developed, aiming to reveal semantically meaningful insights \cite{jin2023guidelines,patricio2023explainable}, which is essential for providing guidance to medical practitioners in making informed decisions or discovering novel biomarkers.

One major class of XAI methods falls under the paradigm of attribution maps. Various methods for computing attribution maps include backpropagation \cite{sundararajan2017axiomatic,xu2020attribution}, activation map \cite{selvaraju2017grad,li2022explainable}, and perturbation-based \cite{zeiler2014visualizing,ribeiro2016should,Petsiuk2018rise}, with the common goal of highlighting the relative importance of each pixel to the model prediction. 
While attribution maps have been relatively useful in identifying salient regions for classification tasks on natural images, their utility has been suboptimal for tasks where different classes share many common features, a situation often seen in image-based prognosis tasks. In these cases, patients may have similar baseline images but different outcomes, and the granularity of traditional attribution maps can hardly pinpoint the human-interpretable differences that lead to different predictions and, therefore, fails to provide the exact reasoning behind their decisions.
To address this issue, Rudin et al. proposed an interpretable model that calculates the similarity of query images to predefined disease-outcome-related prototypes and integrates these similarities into the final decision \cite{rudin2019stop,barnett2021case}. This approach enhances interpretability and reliability substantially by grounding the model's decisions in domain knowledge. However, it requires both extensive prior knowledge and fine annotations, which are not feasible when such knowledge is lacking, and this method may miss predictive features not identified by clinicians. Hence, leveraging model-agnostic explanation methods to detect fine-grained contributive features from deep classifiers offer a promising solution to maintain both interpretability and performance.

Counterfactual explanation, initiated in \cite{wachter2017counterfactual}, became an alternative line of the traditional model-agnostic interpretation methods. Essentially, this approach generates a variant of the original input, \emph{i.e.} the counterfactual, which alters the model's prediction, serving to pinpoint the crucial features that are responsible for this change. Notably, the means for generating the counterfactual examples are greatly facilitated by the parallel developments of the Generative Adversarial Networks (GANs) \cite{goodfellow2014generative,goodfellow2020generative}. 
There has been an emergence of GAN-based counterfactual explanation methods in medical classification tasks across various modalities, including X-Ray \cite{atad2022chexplaining,mertes2022GANterfactual,singla2023explaining,schutte2021using,sankaranarayanan2022real}, Magnetic Resonance Imaging (MRI) \cite{tanyel2023beyond,fontanella2023acat} and ultrasound images \cite{reynaud2022d} and histopathology images \cite{Karras2020ada,schutte2021using}.

While making significant progress in generating realistic counterfactual images, existing counterfactual generation models still face three major limitations that hinder their widespread adoption for explaining various classifiers.
Firstly, training GAN models is notoriously difficult, often requiring meticulous tuning of hyperparameters to ensure stable training \cite{kodali2017convergence,cohen2021gifsplanation,fontanella2023acat}.
Secondly, explaining black-box decisions requires the model to have the ability to reconstruct the original query image without any manipulation applied.
However, GAN-based models often struggle to accurately reconstruct original images with rich textures, resulting in ineffective counterfactuals where only certain features are altered from the original images with more complex contents \cite{schutte2021using} and impeding their adoption in more complex modalities such as histopathology and CT imaging.
Thirdly, most counterfactual generation models require training with inputs from the black-box classifiers they aim to explain \cite{atad2022chexplaining}. This requirement significantly increases training costs and complicates the application of these methods.

In this work, we develop a novel counterfactual method to address the aforementioned limitations, leveraging diffusion-based generation \cite{ho2020denoising,song2020denoising} which has surpassed GANs for many generative tasks \cite{dhariwal2021diffusion,muller2022diffusion}. We aim to achieve a more stable training procedure, higher-quality, controllable counterfactual image generation performance and faster development of explanation models . Our main contributions can be summarized as: 
(1). The proposed \textbf{DiffExplainer}, combines teacher-student learning and Diffusion Autoencoders \cite{preechakul2022diffusion} for generating counterfactual images to decode a given black box model.
(2). Compared to traditional attribution maps, DiffExplainer can accurately identify fine-grained features underlying a black box's decision for any given query image. 
Even for instances with indeterminate classification results - typically a significant challenge for existing methods - our method can provide robust and coherent explanations. 
(3). Compared to the existing counterfactual models, we are also the first to develop counterfactual methods specifically for CT modalities and enable the manipulation of lossless reconstructed images. Moreover, DiffExplainer is trained independently of the classifier, allowing for its direct adoption with different classifiers using the same modalities. This feature may facilitate its wider adoption in various tasks in the future.
\section{Method}
\begin{figure}[h]
\begin{center}
\includegraphics[width=0.9\linewidth]
{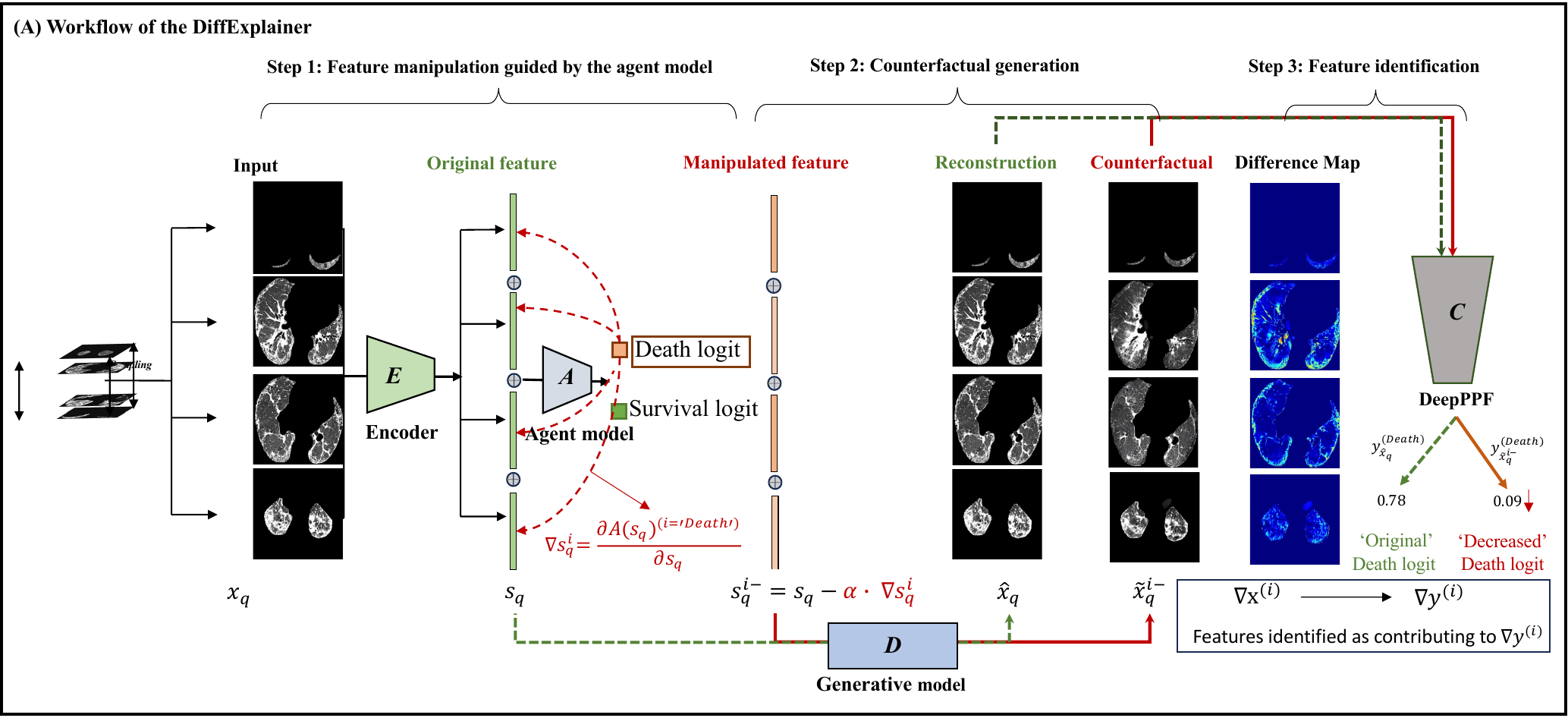}
\includegraphics[width=0.45\linewidth]
{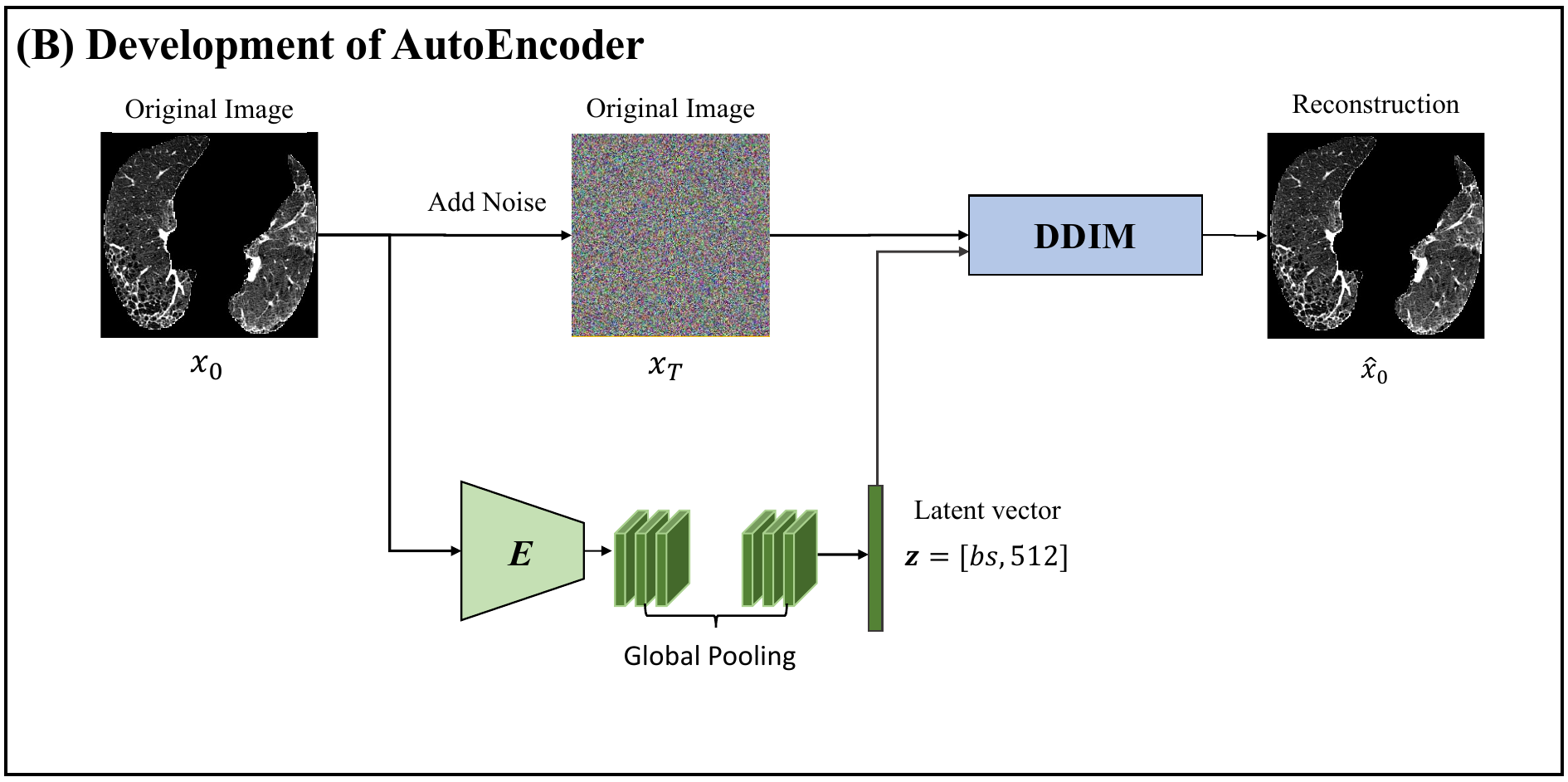}
\includegraphics[width=0.45\linewidth]
{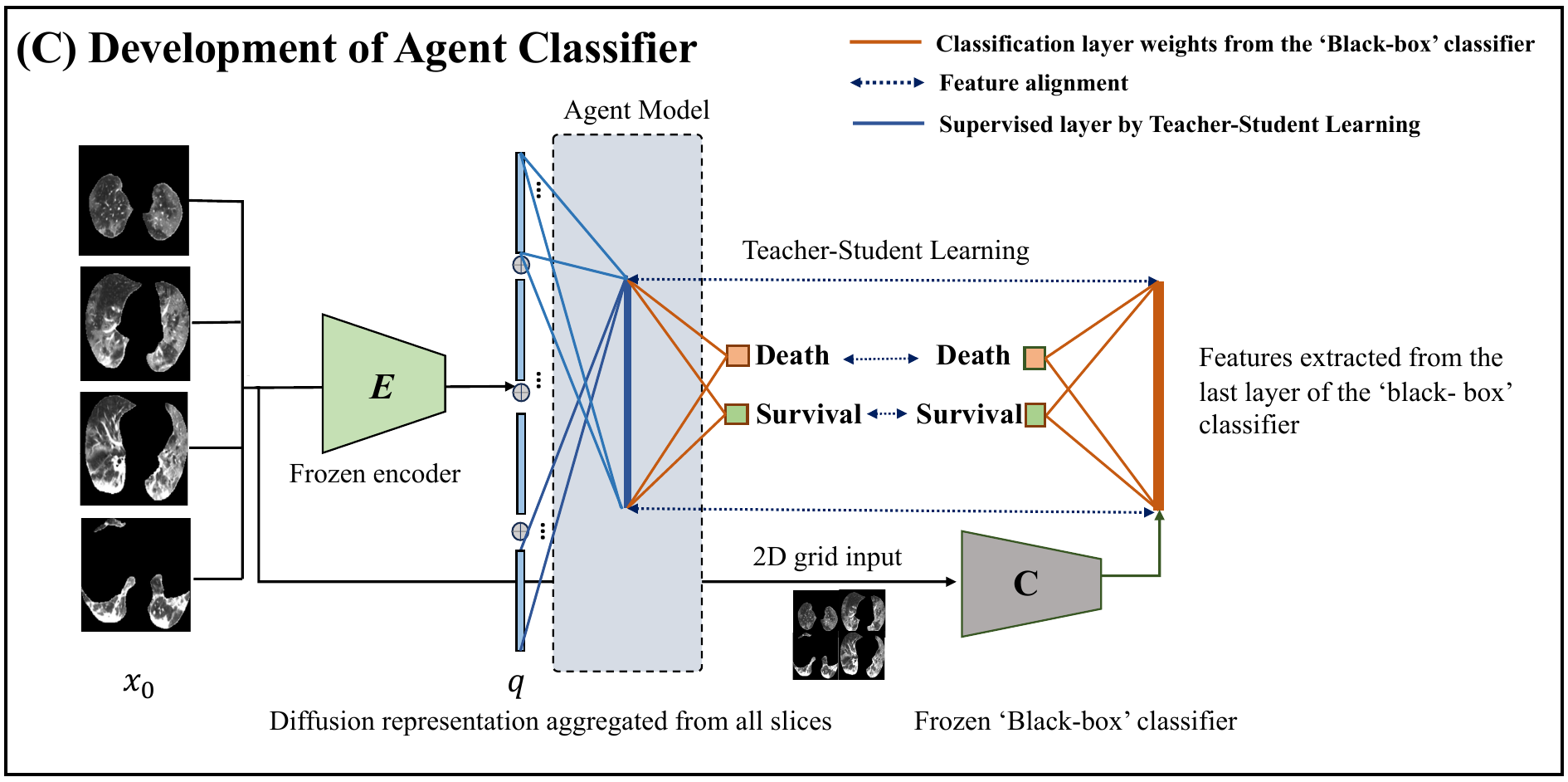}
\end{center}
\caption{ 
Framework of DiffExplainer. (A) The workflow of using DiffExplainer to perform
feature manipulation and counterfactual generation to understand the impact of different regions in affecting the teacher model's predictions.
The two key components are: (B) Diffusion autoencoder consisting of an encoder and a DDIM generative decoder \cite{song2020denoising};
(C) Knowledge distillation for aligning the latent feature from the diffusion autoencoder to that of the given black box.}
\label{fig:framework}
\end{figure}
An overview of our DiffExplainer framework is presented in Fig.~\ref{fig:framework} (A), which showcases elucidating the operations of a black box pretrained classifier for lung disease prognosis. The input of this classifier, which we hereafter refer to as the teacher model, consists of four slices extracted from a CT scan of a patient. Our DiffExplainer, crafted to discern correlated features for model's decisions, comprises three sequentially applied steps with three sub-models.
First, an encoder model maps each slice to a latent representation, which are then combined into a latent feature. Here, the encoder is trained as part of a Diffusion autoencoder model, so that the latent feature therefore contains sufficient information for reconstruction. The feature vector is then sent into an agent model, which performs knowledge distillation and arrives at a final feature representation that is aligned with that of the black box teacher model, hence predicting similar classification scores. This shallow student model is composed of two-layer linear model and will provide the manipulation direction on the latent feature to increase/decrease the score in a target class. In the second stage, we generate counterfactuals from the manipulated features and reconstruct the original images from the original feature from the diffusion model. Lastly, we can generate the difference heatmap from the counterfactual images and the reconstructed images as well as their scores in the black box. The difference maps will give clear indications of the image regions that affect the predictions of the black box. 

\subsection{Diffusion Autoencoder}
A diffusion autoencoder model consists of an encoder that learns a latent representation and a generative decoder that learns to reconstruct the original CT slices. Both the encoder and decoder play crucial roles in the proposed DiffExplainer method and are pretrained jointly, as illustrated in Fig.~\ref{fig:framework} (B).
The encoder $\mathbf{E}$ compresses the input image $\mathbf{x}_0$ into a semantically meaningful low-dimensional representation $\mathbf{z} \in \mathbb{R}^{512}$. This feature space can be manipulated through a subsequently introduced agent model. The decoder then generates the reconstructed slice $\hat{\mathbf{x}}_0$ from $\mathbf{z}$ and the noise injected input $\mathbf{x}_T$ ($T$ iterations of Gaussian noise injection). The reconstruction is achieved by iteratively performing noise removal with a conditional diffusion model $\mathbf{D}_{\theta}$:
\begin{multline}
 \hat{\mathbf{x}}_{t-1}=\sqrt{\alpha_{t-1}}\left(\frac{\hat{\mathbf{x}}_t-\sqrt{1-\alpha_t} \mathbf{D}_\theta\left(\hat{\mathbf{x}}_t, t, \mathbf{z}\right)}{\sqrt{\alpha_t}}\right)\\+\sqrt{1-\alpha_{t-1}} \mathbf{D}_\theta\left(\hat{\mathbf{x}}_t, t, \mathbf{z}\right), \text{with } t=T,T-1,...,1
\end{multline}
where $\alpha_t$ characterizes the variance schedule (for noise injection) \cite{ho2020denoising}, $\hat{\mathbf{x}}_{T}= \mathbf{x}_{T} = \sqrt{\alpha_T} \boldsymbol{x}_0+\sqrt{1-\alpha_T} \boldsymbol{\epsilon}$ and $\boldsymbol{\epsilon} \sim \mathcal{N}(\mathbf{0}, \mathbf{I})$.

The encoder $\mathbf{E}$ and decoder $\mathbf{D}$ are trained concurrently as the way proposed in \cite{preechakul2022diffusion}, optimizing for the noise reconstruction loss with respect to $\theta$ and $\phi$: $\mathcal{L}_{(\phi, \theta)}=\mathbb{E}\left\|\boldsymbol{\epsilon}-\mathbf{D}_\theta\left(\mathbf{x}_t, t, \mathbf{E}_\phi(\mathbf{x}_0)\right)\right\|_1$.
 
\subsection{Teacher-Student Learning for Agent Classifier}
The agent classifier is designed to provide decisions aligned with the pretrained teacher model, with additional interpretability. To achieve this goal, it is implemented as a two-layer linear architecture, taking the latent space of the previously introduced autoencoder as input, as illustrated in Fig. \ref{fig:framework} (C).

The key role of this two-layer agent classifier is to provide controllable manipulation directions for the input features, consistent with the classification scores of the black box teacher model. To align the classification ability of the two models, we transfer knowledge from the pretrained `black-box' classifier to the agent model by teacher-student learning techniques. Specifically, we engage a linear layer to map the encoder extracted features to align with that of the teacher model's features, supervised by a L1 loss. The weight of the classification layer in the agent model is fixed to match that of the teacher model.

\subsection{Counterfactual Generation}
After training the diffusion autoencoder and the agent model, counterfactual images can be generated from a query image. These images are manipulated to potentially modify the prediction scores of the teacher model, providing insights into the reasoning behind the teacher model's decisions.
As illustrated in Fig. \ref{fig:framework} (A), we denote a query image as $I_{q}$, the teacher model as $\mathbf{C}$, and its classification score as $y$, the features extracted from the diffusion model as $q$, the agent model as $\mathbf{A}$, and its corresponding prediction score as $y'=\mathbf{A}(q)$.

The feature $q$ can be readily manipulated to respectively increase or decrease its score in the agent model as:
\begin{align}\label{eqn:counterfactual}
\begin{cases}
q_{i+} = q + \alpha \frac{dy'_i}{dq}, I'_{q_{i+}} = \mathbf{D}(q_{i+}), y_{i+} = \mathbf{C}(I'_{q_{i+}}) \\
q_{i-} = q - \alpha \frac{dy'_i}{dq}, I'_{q_{i-}} = \mathbf{D}(q_{i-}), y_{i-} = \mathbf{C}(I'_{q_{i-}}) \\
\end{cases}
\end{align}
where $\alpha>0$ is the stepsize for controlling the manipulation scale, and $i$ is the target class for score manipulation. $q_{i+}$ (resp. $q_{i-}$) denotes manipulated features with increased (resp. decreased) agent prediction score $y'_{i+}$ (resp. $y'_{i-}$). The corresponding counterfactual images are denoted as $I'_{q_{i+/-}}$, while the reconstructed original image is $I'_q=\mathbf{D}(q)$.

\subsection{Evaluation criteria for counterfactual generation}
We define the following criteria as essential conditions that counterfactual images need to meet in order to effectively explain a black box model: (1) (Reconstruction ability) The generated counterfactual images should retain most of the features present in the query sample ($I_{q_{i\pm}}\approx I_{q}$). When there is no manipulation, the generated counterfactual should have minimal deviations from its query image and similar classification scores. (2) (Observable) The generated counterfactual images should change certain features, that can be detected and discerned by the human eye, while also preserving the overall context of the query image unchanged. (3) (Manipulation) The classification score of the generated counterfactual for the black box should change in the intended manipulation direction, \emph{i.e.} $y_{i\pm}$ should change in the same direction as $y'_{i\pm}$ changes.

\section{Experiments}
\subsection{Dataset}
We trained a DiffExplainer for counterfactual generations for a pretrained classification model that predicts the one-year mortality of patients with Fibrotic Lung Disease (FLD). Specifically, we utilized two FLD datasets. The first cohort is sourced from OSIC \footnote{\url{https://www.osicild.org/}}, with 27 patients deceased within one year and 704 patients surviving for one year. The second cohort is an in-house dataset collected from hospitals in Australia, with 43 patients deceased within one year and 458 patients surviving for one year. In our experiments, we used only the first dataset for training, and switched to the the second in-house dataset for evaluation.

\subsection{Experimental setting}
The black box classification model employs DenseNet121 as backbone, and correspondingly the two-layer Agent model consists of 1024 neurons in the first layer (aligned to the feature dimension of the black box) and two neurons in the second layer for the two-classification task. For training the black box and agent model, we partitioned this dataset into 5 folds for training and selected the models with the highest sum of specificity and sensitivity. Both models are trained with a batch size of 16 for 100 epochs, with learning rates of 1e-6 and $1e^{-4}$, respectively. The autoencoder, with an input size of 256$\times$256, was trained on slices from the OSIC dataset using eight V100 GPUs with a learning rate of $1e^{-4}$ and a batch size of 64 for 100 epochs.

\subsection{Validity of the Agent Model}
We first evaluate the consistency of the agent model in producing similar predictions as its black box teacher model. To proceed, we designate the predictions provided by the black box as the `ground truth'. We calculate the accuracy, sensitivity, specificity, and F1 score for the agent model's results with respect to this `ground truth' on the test dataset, to validate if the agent model is indeed aligned with its teacher model. Furthermore, we report the Kullback–Leibler (KL) divergence for the two models' predicted probability. We also conducted ablation experiments to study how different training objectives affects knowledge distillation and feature alignment, including L2 and L1 loss on feature space and KL divergence on the model's output. The alignment performance under different constraints is presented in Table \ref{tab:agentperformance}. It can be seen that these different training objectives are all effective in ensuring feature alignment.
\begin{table}
\centering
\caption{Alignment of agent model to teacher model.}\label{tab:agentperformance}
\begin{tabular}{c|cccccc}
\hline
Method & AUC & Accuracy & Sensitivity & Specificity & F1 & KL \\ \hline
MSE & 0.99 & 0.95 & 0.91 & 0.97 & 0.92 & 0.18 \\ \hline
L1 & 0.99 & 0.96 & 0.91 & 0.98 & 0.93 & 0.17 \\ \hline
KL & 0.99 & 0.96 & 0.91 & 0.98 & 0.93 & 0.17 \\ 
\hline
\end{tabular}
\end{table}
\subsection{Performance of the counterfactual generation}
\paragraph{The Diffusion autoencoder achieves the best reconstruction quality while GAN-based models fails to achieve acceptable reconstruction.}
The reconstruction from our Diffusion autoencoder achieved an average PSNR of 35.28, SSIM of 0.99, and a KL divergence for classification, demonstrating outstanding reconstruction capability. On the other hand, although capable of synthesizing realistic-looking CT images, the conditional GAN and StyleGAN-based autoencoder models in our experiments failed to accurately reconstruct the same textures such as airway and vessel from the encoded features of the query image (this is also observed in the histology reconstruction task in \cite{schutte2021using}).Since GAN based autoencoders failed the reconstruction criterion, we did not continue to generate counterfactual examples.

\begin{figure}[ht]
\begin{subfigure}{\textwidth}
\hspace{0.08\textwidth}
\begin{minipage}[t]{0.16\textwidth}
\centering
\parbox{2cm}{\raggedright\fontsize{3}{1}\selectfont Original:\\ \textbf{Death: 0.25}\\Survival: 0.02}
\includegraphics[width=\linewidth]{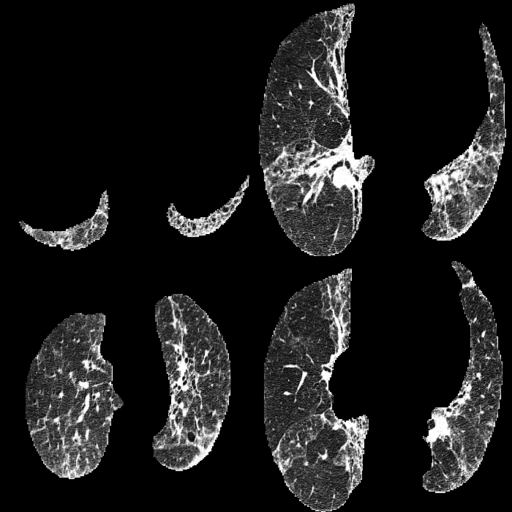}
\includegraphics[width=\linewidth]{Figures/00620_orig.png}
\end{minipage}
\begin{minipage}[t]{0.16\linewidth}
\centering
\parbox{2cm}{\fontsize{3}{1}\selectfont$\alpha=+10$:\\ 
\textbf{Death: 0.67} \\Survival: -0.24}
\includegraphics[width=\linewidth]{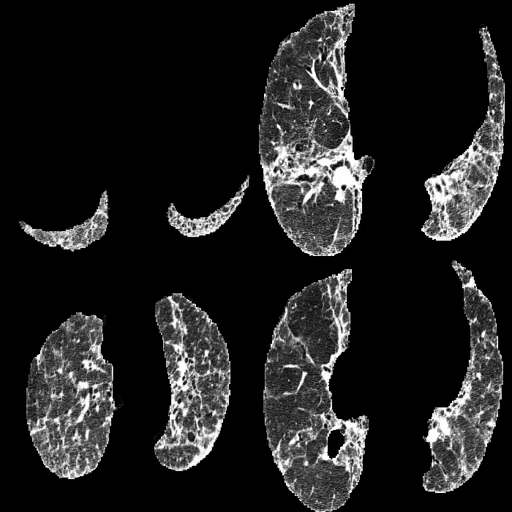}
\includegraphics[width=\linewidth]{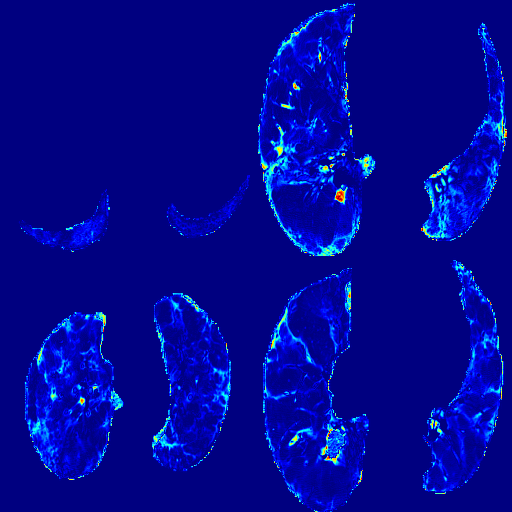}
\end{minipage}
\begin{minipage}[t]{0.16\linewidth}
\centering
\parbox{2cm}{\fontsize{3}{1}\selectfont$\alpha=+15$:\\ 
\textbf{Death: 0.81} \\Survival: -0.29}
\includegraphics[width=\linewidth]{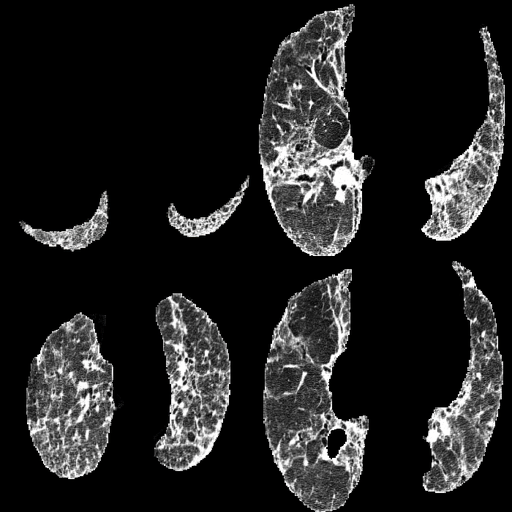}
\includegraphics[width=\linewidth,height=\linewidth]{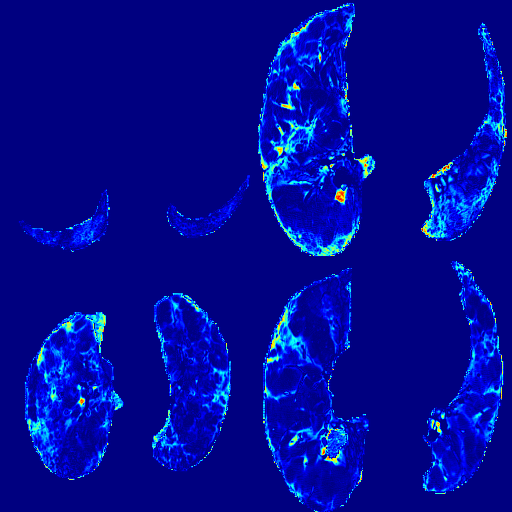}
\end{minipage}
\begin{minipage}[t]{0.16\linewidth}
\centering
\parbox{2cm}{\fontsize{3}{1}\selectfont$\alpha=+20$:\\ 
\textbf{Death: 0.88} \\Survival: -0.33}
\includegraphics[width=\linewidth]{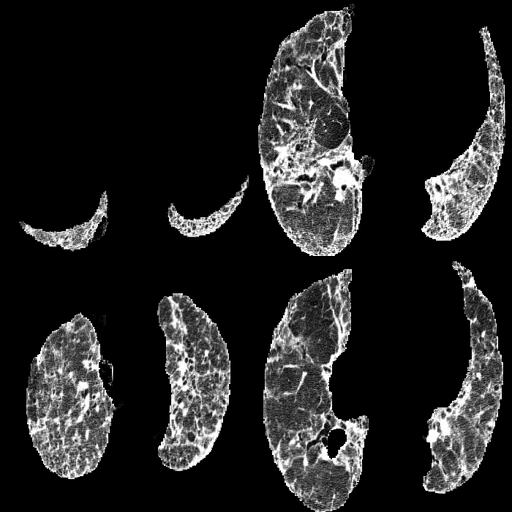}
\includegraphics[width=\linewidth,height=\linewidth]{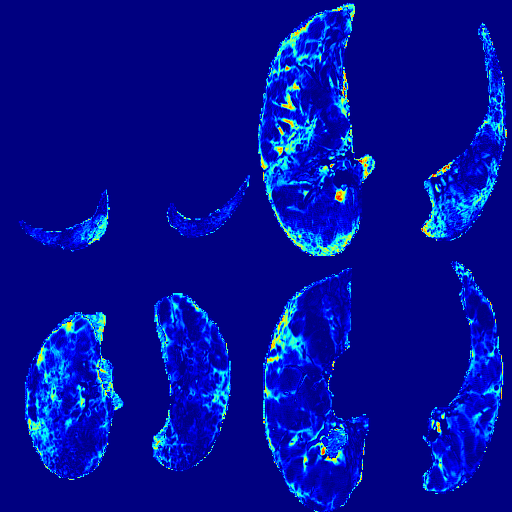}
\end{minipage}
\begin{minipage}[t]{0.16\linewidth}
\centering
\parbox{2cm}{\fontsize{3}{1}\selectfont$\alpha=+30$:\\ 
\textbf{Death: 0.90} \\Survival: -0.34}
\includegraphics[width=\linewidth]{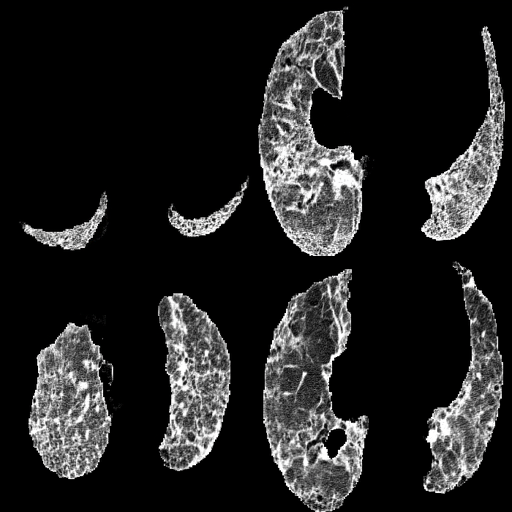}
\includegraphics[width=\linewidth,height=\linewidth]{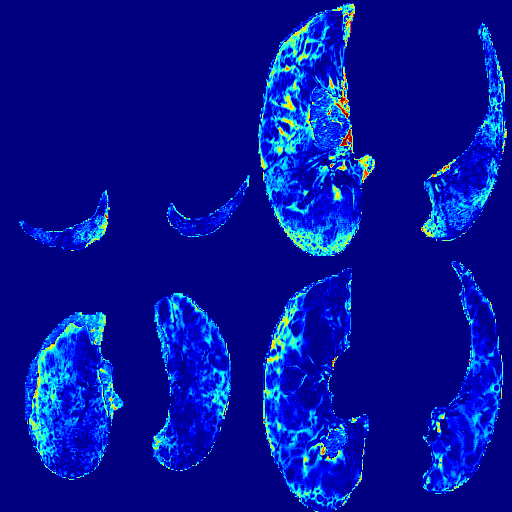}
\end{minipage}
\end{subfigure}
\begin{subfigure}{\textwidth}
\hspace{0.08\textwidth}
\begin{minipage}[t]{0.16\linewidth}
\centering
\parbox{2cm}{\raggedright\fontsize{3}{1}\selectfont Reconstructed:\\ Death: 0.24\\ \textbf{Survival: 0.10}}
\includegraphics[width=\linewidth]{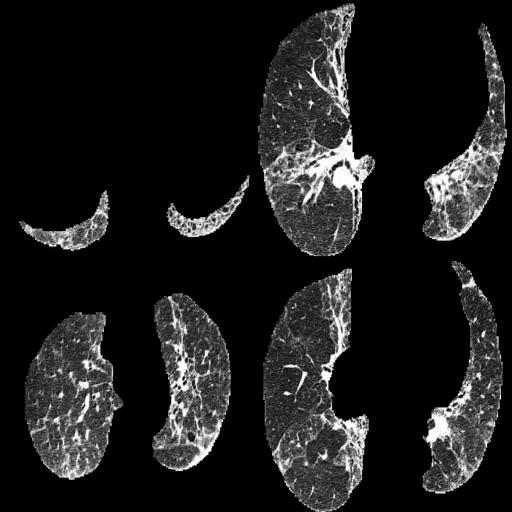}
\includegraphics[width=\linewidth]{Figures/00620_recon.png}
\end{minipage}
\begin{minipage}[t]{0.16\linewidth}
\centering
\parbox{2cm}{\fontsize{3}{1}\selectfont$\alpha=-10$:\\ 
Death: -0.08, \\ \textbf{Survival: 0.42}}
\includegraphics[width=\linewidth]{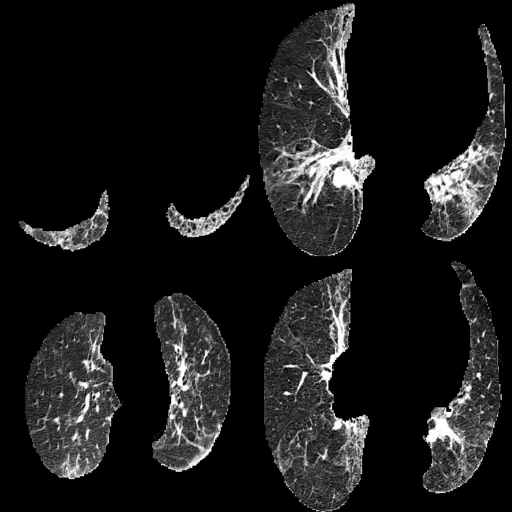}
\includegraphics[width=\linewidth,height=\linewidth]{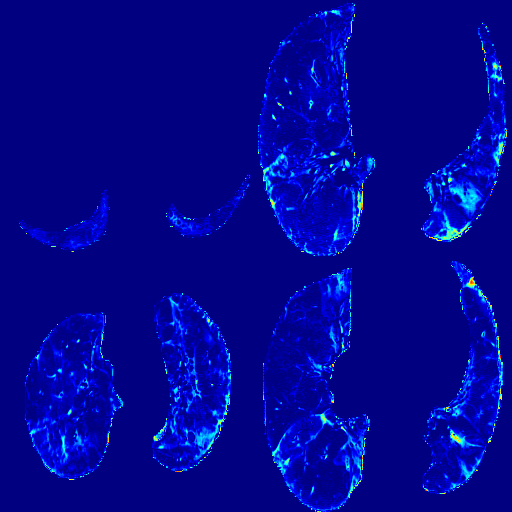}
\end{minipage}
\begin{minipage}[t]{0.16\linewidth}
\centering
\parbox{2cm}{\fontsize{3}{1}\selectfont$\alpha=-15$:\\ 
Death: -0.25, \\
\textbf{Survival: 0.61}}
\includegraphics[width=\linewidth]{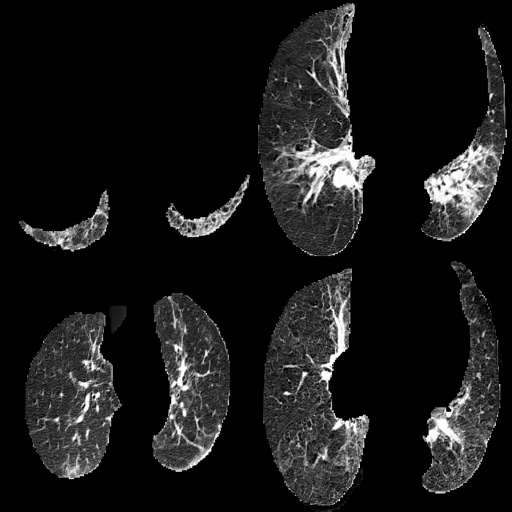}
\includegraphics[width=\linewidth,height=\linewidth]{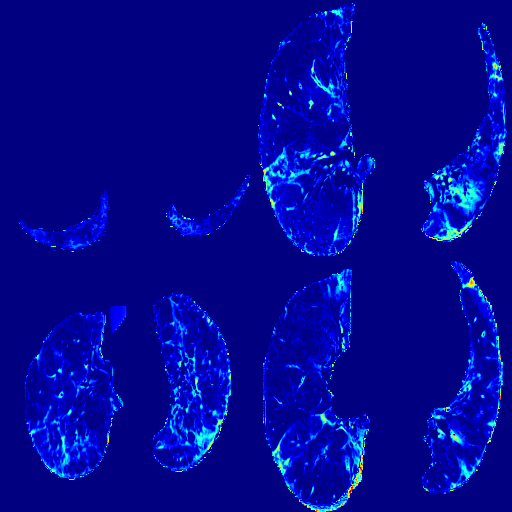}
\end{minipage}
\begin{minipage}[t]{0.16\linewidth}
\centering
\parbox{2cm}{\fontsize{3}{1}\selectfont$\alpha=-20$:\\ 
Death: 0.22, \\ \textbf{Survival: 0.78}}
\includegraphics[width=\linewidth]{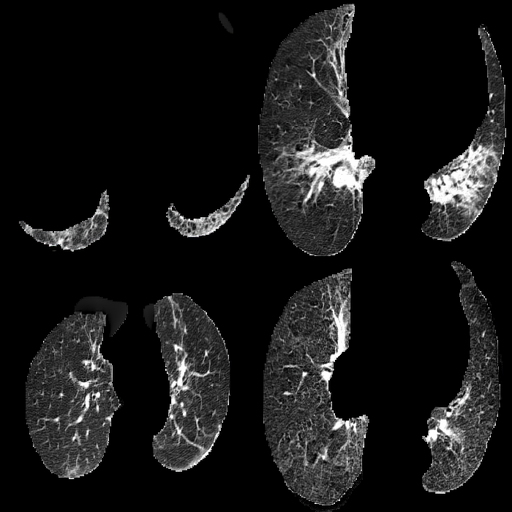}
\includegraphics[width=\linewidth,height=\linewidth]{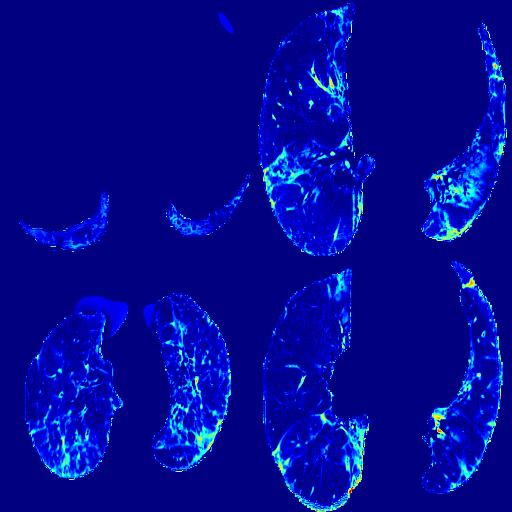}
\end{minipage}
\begin{minipage}[t]{0.16\linewidth}
\centering
\parbox{2cm}{\fontsize{3}{1}\selectfont$\alpha=-30$:\\ 
Death: 0.12, \\ \textbf{Survival: 0.88}}
\includegraphics[width=\linewidth]{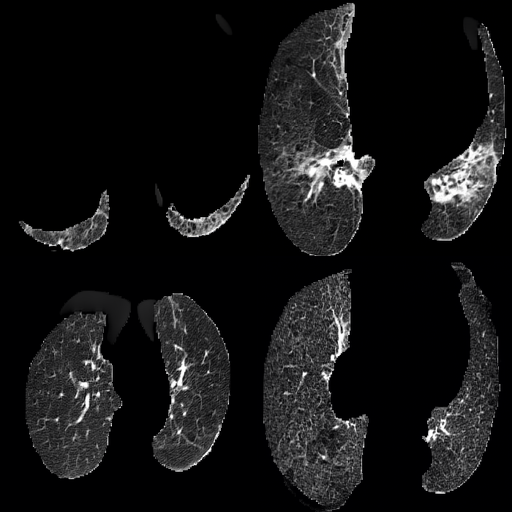}
\includegraphics[width=\linewidth,height=\linewidth]{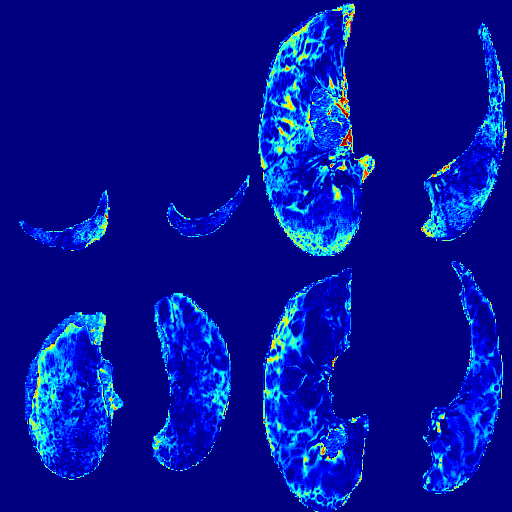}
\end{minipage}
\end{subfigure}
\caption{Counterfactual generation for `hard' cases that the pretrained classifier failed to assign confident predictions. Row 1: Counterfactuals with increasing `Death score'; Row 2: Counterfactuals with increasing `Survival score'.}
\label{fig:synimg}
\end{figure}

\paragraph{DiffExplainer consistently pinpoints observable features upon which the decision is based.}
We select the cases which are indeterminate to the ‘black box’ (similar logits for each class) and manipulate it to achieve a confident score for each class. As presented in Fig. \ref{fig:synimg}, we generated the counterfactual examples which can be confidently classified as the 1-year mortality and 1-year survival. The difference between the counterfactuals and the original images are presented as difference heatmaps, where red indicates the increased textures and blue indicate the decreased textures. We can attribute the indeterminacy of these cases to the lack of features which are leveraged by the ‘black-box’ classifier.

\paragraph{DiffExplainer allows for fine-grained control over the counterfactual generation, enabling smooth transition from one classification result to another.}
In Fig. \ref{fig:synimg}, we illustrate the controllability of the counterfactual generation method by incrementally increasing the manipulation weight. The resulting generated images were validated to exhibit a gradual augmentation in the decisions of the black box classifier and manifestation of features. 

\subsection{Comparison to other XAI methods}
We compare the difference heatmaps obtained by DiffExplainer approach to the attribution maps from existing widely-employed XAI methods. 
We observe that DiffExplainer can more accurately pinpoint theregions within the query image influencing model decisions, with finer granularity, whereas other methods can only localize to a coarse area (Grad-CAM methods) or to regions that are incomprehensible (Saliency maps). For an indeterminate case, DiffExplainer is capable of generating the missing features that can lead to confident decisions for either class, thus explaining why the model arrived at its indeterminate decision, a capability not matched by existing methods.

\begin{figure}[h]
\begin{minipage}[t]{0.135\linewidth}
\centering
\parbox{2cm}{\raggedright\fontsize{3}{1}\selectfont Original:\\ \textbf{Death: 0.10}\\Survival: 0.11}
\includegraphics[width=\linewidth]
{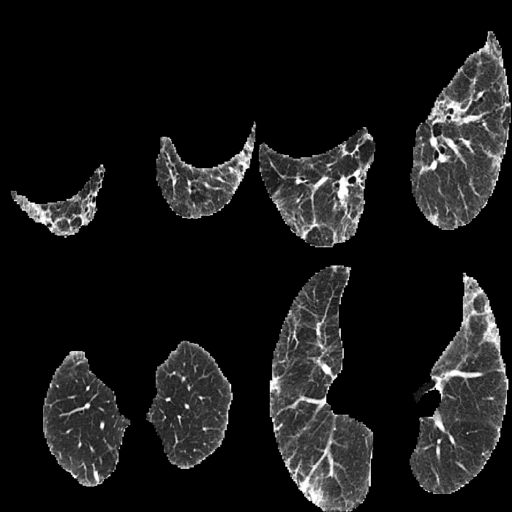}
\parbox{2cm}{\raggedright\fontsize{3}{1}\selectfont Reconstructed:\\ \textbf{Death: 0.12}\\Survival: 0.12}
\includegraphics[width=\linewidth]{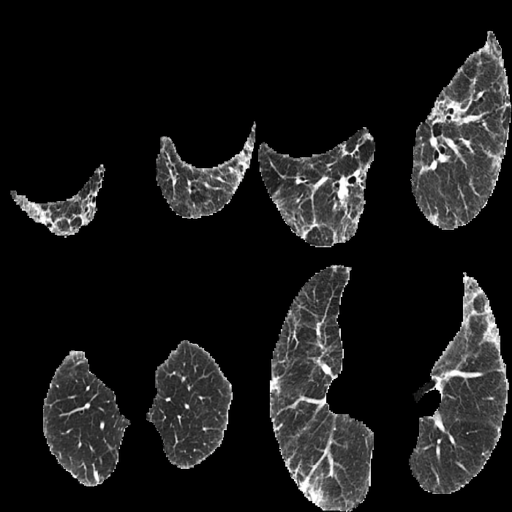}
\end{minipage}
\begin{minipage}[t]{0.135\linewidth}
\centering
\parbox{2cm}{\fontsize{3}{1}\selectfont$\alpha=-15$:\\ 
\textbf{Death: -0.39} \\Survival: 0.63}
\includegraphics[width=\linewidth, height=\linewidth]
{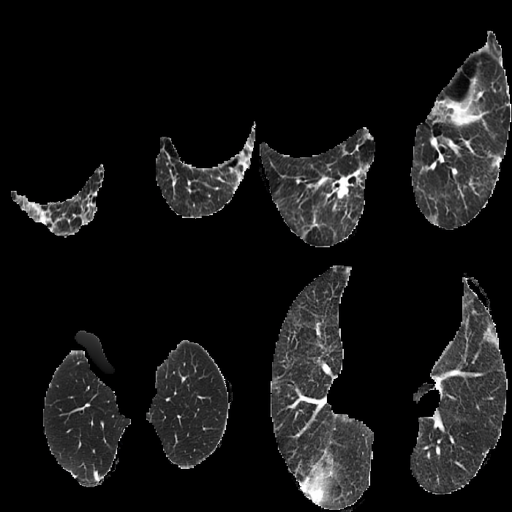} 
\parbox{2cm}{\fontsize{3}{1}\selectfont$\alpha=+15$:\\ 
\textbf{Death: 0.76} \\Survival: -0.37}
\includegraphics[width=\linewidth]{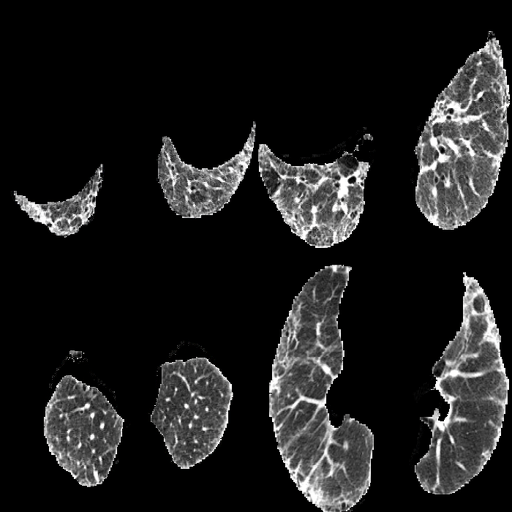}
\end{minipage}
\begin{minipage}[t]{0.135\linewidth}
\centering
\parbox{2cm}{\fontsize{3}{1}\selectfont
Existing feature\\`Death' $\uparrow$\vspace{.5em}}
\includegraphics[width=\linewidth, height=\linewidth]
{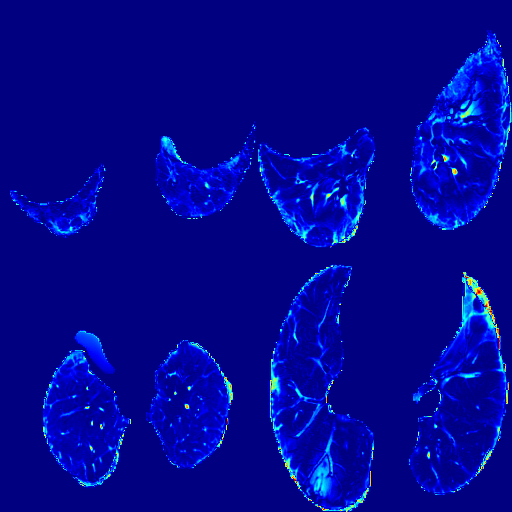} 
\parbox{2cm}{\fontsize{3}{1}\selectfont
Missing feature\\`Survival' $\uparrow$\vspace{.5em}}
\includegraphics[width=\linewidth,height=\linewidth]{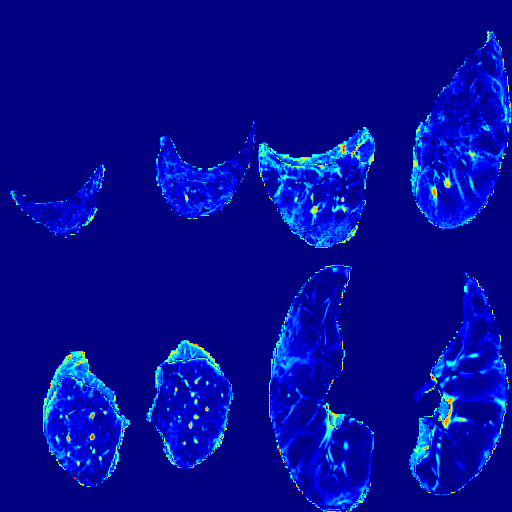}
\end{minipage}
\begin{minipage}[t]{0.135\linewidth}
\centering
\parbox{2cm}{\fontsize{3}{1}\selectfont 
GradCAM\cite{selvaraju2017grad}\vspace{1.3em}}
\includegraphics[width=\linewidth, height=\linewidth]{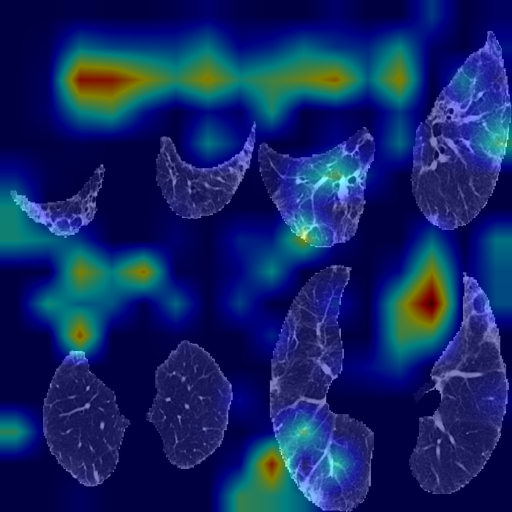}
\parbox{2cm}{\fontsize{3}{1}\selectfont 
GradCAM\cite{selvaraju2017grad}\vspace{1.3em}}
\includegraphics[width=\linewidth,height=\linewidth]{Figures/00438_recon_gradcam1.png}
\end{minipage}
\begin{minipage}[t]{0.135\linewidth}
\centering
\parbox{2cm}{\fontsize{3}{1}\selectfont 
GradCAM++\cite{chattopadhay2018grad}\vspace{1.3em}}
\includegraphics[width=\linewidth, height=\linewidth]{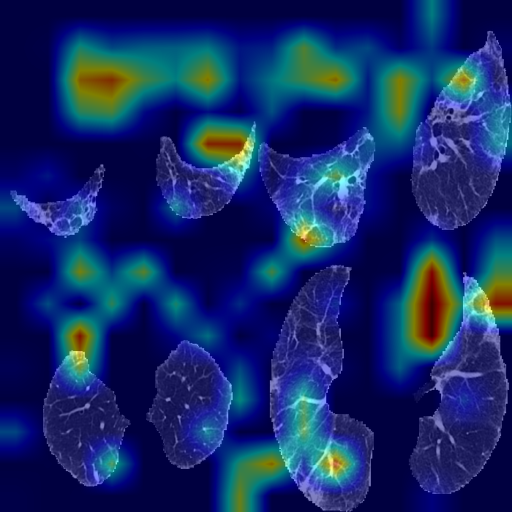}
\parbox{2cm}{\fontsize{3}{1}\selectfont 
GradCAM++\cite{chattopadhay2018grad}\vspace{1.3em}}
\includegraphics[width=\linewidth,height=\linewidth]{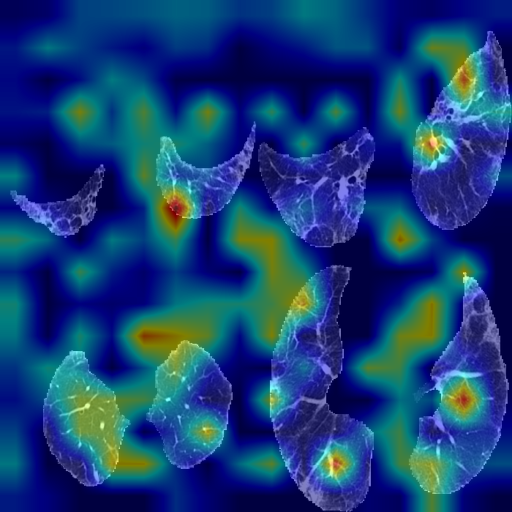}
\end{minipage}
\begin{minipage}[t]{0.135\linewidth}
\centering
\parbox{2cm}{\fontsize{3}{1}\selectfont
Blur IG\cite{xu2020attribution}\vspace{1.3em}}
\includegraphics[width=\linewidth, height=\linewidth]{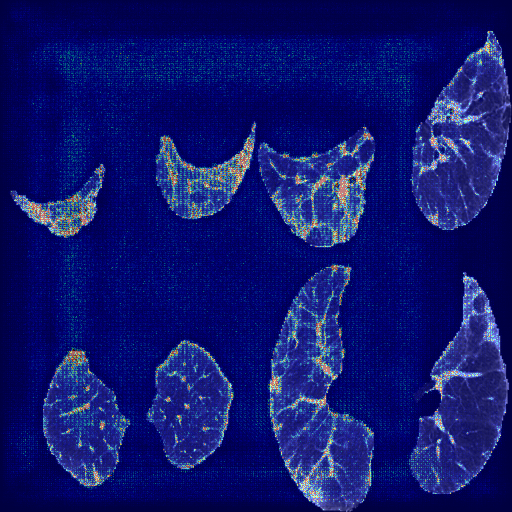}
\parbox{2cm}{\fontsize{3}{1}\selectfont
Blur IG\cite{xu2020attribution}\vspace{1.3em}}
\includegraphics[width=\linewidth,height=\linewidth]{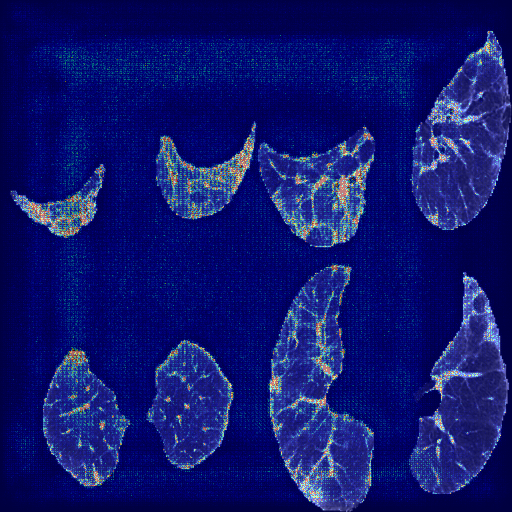}
\end{minipage}
\begin{minipage}[t]{0.135\linewidth}
\centering
\parbox{2cm}{\fontsize{3}{1}\selectfont
XRAI\cite{kapishnikov2019xrai}\vspace{1.3em}}
\includegraphics[width=\linewidth, height=\linewidth]{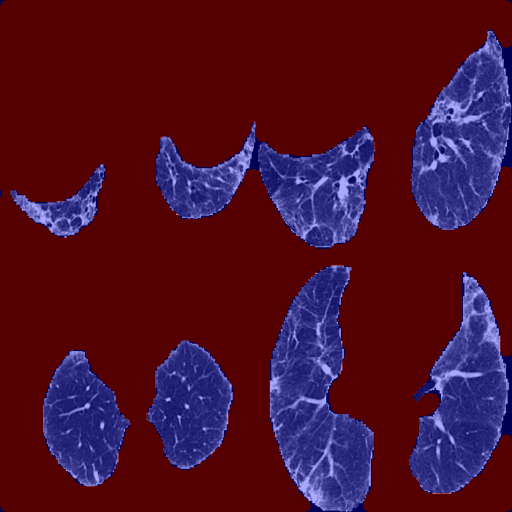}
\parbox{2cm}{\fontsize{3}{1}\selectfont 
XRAI\cite{kapishnikov2019xrai}\vspace{1.3em}}
\includegraphics[width=\linewidth,height=\linewidth]{Figures/00438_recon_xrai0.png}
\end{minipage}
\caption{Comparison to other XAI methods. Blue and red areas indicate the existing features and missing features that contributes to the change in prediction.}
\label{synimg}
\centering
\end{figure}

\section{Conclusion}
In this paper, we developed a counterfactual generation approach named DiffExplainer for explaining the decisions of a given black box classifier. Compared to existing attribution methods, our approach can more accurately locate the features influencing model decisions within the query image, has finer granularity, and also allows enhanced controllability. We validate our approach on the challenging modality of CT scans, which posed significant difficulties for existing methods. Our DiffExplainer distinguishes itself by providing high-quality reconstructions, and the capacity to identify semantically meaningful features which impact the classifier's predictions. This capability enables users to uncover the reasons behind each decision made by a given black box model. Combining DiffExplainer with highly accurate black-box models presents an opportunity to discover novel biomarkers.

\subsubsection{\ackname} This study was supported in part by the ERC IMI (101005122), the H2020 (952172), the MRC (MC/PC/21013), the Royal Society (IEC/NSFC/211235), the NVIDIA Academic Hardware Grant Program, the SABER project supported by Boehringer Ingelheim Ltd, Wellcome Leap Dynamic Resilience, NIHR Imperial Biomedical Research Centre, and the UKRI Future Leaders Fellowship(MR/V023799/1).

\subsubsection{\discintname}
The authors have no competing interests to declare that are relevant to the content of this article. 

\bibliographystyle{splncs04}
\bibliography{reference}
\end{document}